\documentclass[11pt]{article}

\usepackage[final]{acl}

\usepackage{times}
\usepackage{latexsym}

\usepackage[T1]{fontenc}

\usepackage[utf8]{inputenc}

\usepackage{microtype}

\usepackage{inconsolata}

\usepackage{graphicx}

\usepackage{amsmath}
\usepackage{amsfonts}
\usepackage{amssymb}
\usepackage{dblfloatfix}
\usepackage{xcolor}
\usepackage{booktabs}
\usepackage{colortbl}
\usepackage{multirow}
\title{Hallucinations as Orthogonal Noise: Inference-Time Manifold Alignment via Dynamic Contextual Orthogonalization}

\author{%
	\textbf{Mingkuan Zhao\textsuperscript{1}}\thanks{These authors contributed equally to this work.},
	\textbf{Wentao Hu\textsuperscript{1,2}}\footnotemark[1],
	\textbf{Tianchen Huang\textsuperscript{4}},
	\textbf{Yuheng Min\textsuperscript{5}},
	\textbf{Suquan Chen\textsuperscript{1}},
	\\
	\textbf{Yide Gao\textsuperscript{1}},
	\textbf{Yanbo Zhai\textsuperscript{1}},
	\textbf{Shuangyong Song\textsuperscript{2}},
	\textbf{Xuelong Li\textsuperscript{3}}\thanks{Corresponding author.}
	\\[0.8em]
	\parbox{\linewidth}{\small\centering
		\textsuperscript{1}Xi'an Jiaotong University,
		\textsuperscript{2}Xingchen AGI Lab, China Telecom AI Technology (Beijing) Co., Ltd.,\\
		\textsuperscript{3}Institute of Artificial Intelligence, China Telecom,
		\textsuperscript{4}University of Science and Technology of China,
		\textsuperscript{5}Tsinghua University
	}
	\\[0.6em]
	\parbox{\linewidth}{\small\centering
		\texttt{\{mingkuanzhao, wentao\_hu, yanbozhai, suquanchen, yidegao\}@stu.xjtu.edu.cn},\\
		\texttt{tchuang@mail.ustc.edu.cn},
		\texttt{minyh24@mails.tsinghua.edu.cn},\\
		\texttt{songshy@chinatelecom.cn},
		\texttt{xuelong\_li@ieee.org}
	}
}

\begin{document}
\maketitle
\begin{abstract}
	Hallucination in Large Language Models (LLMs)—characterized by the generation of content inconsistent with contextual facts or logical constraints—remains a persistent challenge for reliable deployment. In this work, we address this issue through a geometric framework rooted in the linear representation hypothesis. We propose that hallucinations manifest as \textit{orthogonal noise} relative to the semantic manifold of the residual stream. Specifically, we hypothesize that while attention heads ideally propagate information congruent with the context subspace, hallucinations arise when specific heads introduce components orthogonal to this subspace, disrupting the coherence of the latent representation.
	
	Based on this formulation, we introduce \textbf{Dynamic Contextual Orthogonalization (DCO)}, an inference-time intervention method. DCO utilizes the input residual stream as a dynamic context anchor to perform orthogonal decomposition on attention head outputs. To distinguish between context-aligned semantic updates and divergent noise, DCO employs a layer-wise Z-score suppression mechanism that selectively attenuates outlier orthogonal components based on statistical distributions. 
	
	Evaluations on Llama-3-8B and 70B across benchmarks such as XSum, NQ-Swap, and IFEval demonstrate that DCO achieves superior contextual faithfulness compared to state-of-the-art intervention baselines. Furthermore, DCO maintains high performance on knowledge-intensive tasks like TriviaQA and TruthfulQA, effectively mitigating the trade-off between hallucination suppression and parametric knowledge retention often observed in existing methods. Our findings validate the geometric interpretation of hallucinations and establish DCO as a computationally efficient approach for enforcing manifold alignment. Our code is available at \url{https://github.com/Harry-Miral/DCO}.
\end{abstract}

\section{Introduction}

The deployment of Large Language Models (LLMs) in critical domains is hindered by the persistence of hallucinations, where generated content diverges from contextual facts or logical constraints. While current mitigation strategies such as Retrieval-Augmented Generation (RAG) or contrastive decoding effectively modulate output probabilities, they often operate without a mechanistic view of the internal representation dynamics. We posit that hallucinations are not merely surface-level statistical anomalies but manifestations of geometric misalignment within the model's latent space, specifically when the model processes conflicting or ambiguous contexts.

This study approaches the hallucination problem through the \textit{Linear Representation Hypothesis} \citep{park2024linearrepresentationhypothesisgeometry}, which suggests that semantic concepts are encoded as linear directions in high-dimensional space \citep{gurnee2024languagemodelsrepresentspace}. Under this framework, we formalize the faithfulness of generation as a manifold alignment task. Ideally, attention heads should propagate information that lies parallel to the subspace defined by the context. We hypothesize that hallucinations arise from \textit{Orthogonal Noise Injection}—a phenomenon where specific attention heads introduce components orthogonal to the established context manifold, thereby steering the latent state evolution away from the factual trajectory.

\begin{figure}[ht]
	\centering
	\includegraphics[width=1.0\columnwidth]{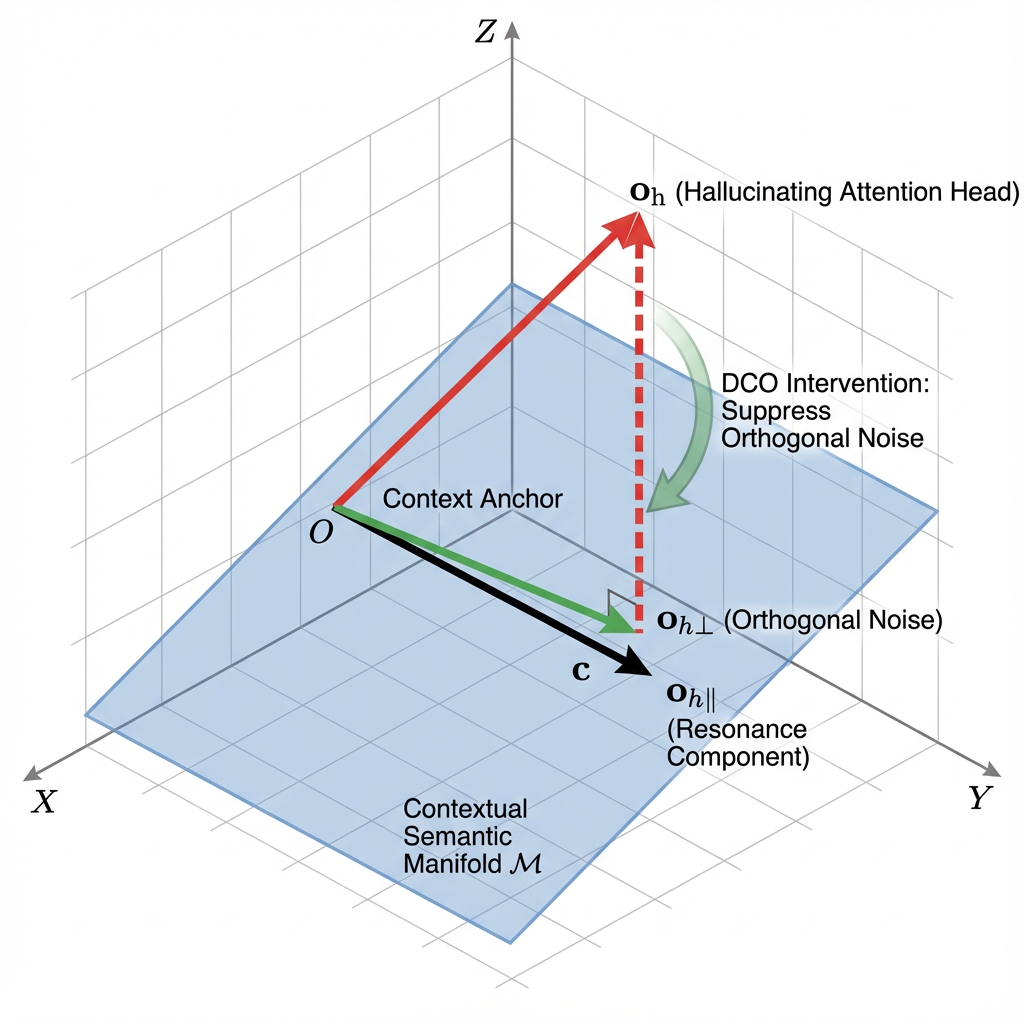}
	\caption{\textbf{Conceptual geometry of Dynamic Contextual Orthogonalization (DCO).} 
		A hallucinating attention head produces an output vector $\mathbf{o}_h$ that diverges from the contextual semantic manifold $\mathcal{M}$. 
		DCO decomposes $\mathbf{o}_h$ into a context-aligned resonance component $\mathbf{o}_{h\parallel}$ and an orthogonal noise component $\mathbf{o}_{h\perp}$. 
		By dynamically attenuating $\mathbf{o}_{h\perp}$ based on layer-wise statistical distributions, DCO realigns the latent representation with the established manifold.}
	\label{fig:teaser}
\end{figure}

Based on this geometric formulation, we introduce \textbf{Dynamic Contextual Orthogonalization (DCO)}, an inference-time intervention method. Unlike static steering vectors that apply a fixed direction, DCO utilizes the input residual stream of each layer as a dynamic \textit{context anchor}. The method performs orthogonal decomposition on attention head outputs to isolate components that deviate from the global semantic consensus. Crucially, to distinguish between necessary information diversification and harmful noise, DCO employs a layer-wise Z-Score suppression mechanism. This adaptive filtering is derived from the observation that hallucinatory signals manifest as statistical outliers in the orthogonality distribution, necessitating a dynamic rather than static threshold to preserve the model's parametric knowledge retrieval capabilities.

We evaluate DCO on the Llama-3-8B and 70B models across a diverse set of benchmarks. Empirical results indicate that DCO achieves superior performance compared to state-of-the-art intervention methods, including Inference-Time Intervention (ITI) and Decoding by Contrasting Layers (DoLa). Specifically, DCO significantly enhances contextual faithfulness on XSum and IFEval while maintaining robustness on knowledge-intensive tasks such as TriviaQA and MuSiQue. These findings demonstrate that enforcing geometric constraints in the residual stream offers a computationally efficient and mechanistically grounded approach to hallucination mitigation, achieving an optimal trade-off between faithfulness and general capability.

\section{Related Work}
\begin{figure*}[t]
	\centering
	\includegraphics[width=0.8\textwidth]{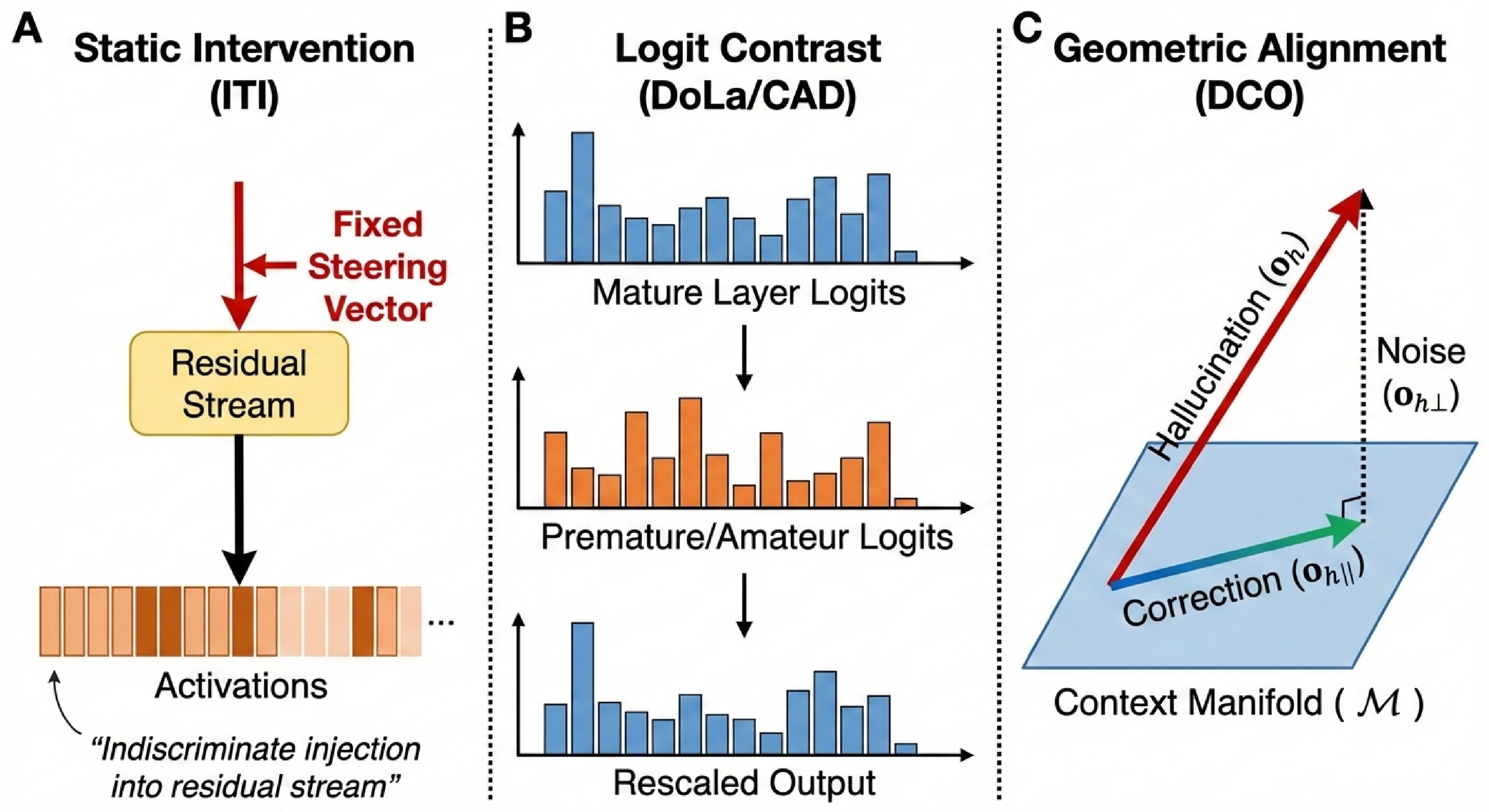} 
	\caption{\textbf{Mechanistic comparison of intervention paradigms.} 
		\textbf{(A) Static Intervention (ITI):} Injects a fixed steering vector derived from offline probing, applying indiscriminate directionality to the residual stream regardless of context.
		\textbf{(B) Logit Contrast (DoLa/CAD):} Operates at the output layer by subtracting premature/amateur logits from mature layer logits, enhancing sharpness but failing to correct internal representation dynamics.
		\textbf{(C) Geometric Alignment (DCO, Ours):} Performs dynamic intervention within the latent space. It treats hallucination as an orthogonal noise component $\mathbf{o}_{h\perp}$ diverging from the context manifold $\mathcal{M}$ and projects the attention output $\mathbf{o}_h$ to align with the valid semantic subspace $\mathbf{o}_{h\parallel}$.}
	\label{fig:comparison}
\end{figure*}

\paragraph{Mechanistic Interpretability of Hallucinations.}
Understanding the computational underpinnings of Transformer models is essential for diagnosing generation failures. Grounded in the view of the residual stream as a communication channel modulated by independent attention heads \citep{dar-etal-2023-analyzing}, mechanistic interpretability has identified specific components governing information flow. Key findings include the characterization of induction heads \citep{olsson2022incontextlearninginductionheads} and ``retrieval heads'' that extract factual content from long contexts \citep{wu2024retrievalheadmechanisticallyexplains}. Moreover, research on knowledge localization \citep{10.5555/3600270.3601532, geva-etal-2023-dissecting} elucidates how factual associations are retrieved via the interplay between MLP layers and attention mechanisms. Recent work by \citet{nanda2023progressmeasuresgrokkingmechanistic} and \citet{yu-etal-2024-mechanistic} further distinguishes between early-layer knowledge gaps and upper-layer retrieval failures as distinct causes of hallucination. Complementary to this, \citet{zhao2025makingheadcountsparse} demonstrate that attention heads exhibit heterogeneous importance, showing that sparse attention patterns can be enforced without a speed-performance trade-off—an observation that supports our hypothesis of differential head behavior in the context of hallucinatory noise. Our work operationalizes these insights, translating the theoretical understanding of attention head dynamics into a direct intervention method within the residual stream.

\paragraph{Inference-Time Hallucination Mitigation.}
Various training-free decoding strategies have been developed to intervene during inference. Methods such as Contrastive Decoding (CD) \citep{li-etal-2023-contrastive} and its variants, including DoLa \citep{chuang2023dola} and Autocontrastive Decoding \citep{gera-etal-2023-benefits}, adjust next-token probabilities by contrasting distributions across model layers or scales. Addressing contextual faithfulness specifically, \citet{shi2024trusting} proposed Context-Aware Decoding (CAD) to amplify evidence-based logits, while \citet{chen2024context} utilized in-context sharpness as a detection signal. More recently, \citet{gema2024decore} introduced DeCoRe to mask retrieval heads based on static importance scores. Crucially, these approaches predominantly operate at the output logit level or employ static interventions \textbf{(as illustrated in Figure \ref{fig:comparison})}. In contrast, guided by the \textit{Linear Representation Hypothesis} \citep{park2024linearrepresentationhypothesisgeometry, hernandez2024linearityrelationdecodingtransformer}, DCO implements a \textit{dynamic} orthogonalization mechanism. Unlike Inference-Time Intervention (ITI) \citep{li2024inferencetimeinterventionelicitingtruthful}, which applies fixed steering vectors derived from offline probing, DCO dynamically calibrates the intervention based on the instantaneous geometry of the residual stream at each step.

\paragraph{Geometric Perspectives on Semantic Consistency.}
We conceptualize generative faithfulness as a manifold alignment problem. In contexts involving knowledge conflicts or long-range dependencies, parametric memory often interferes with contextual extraction \citep{longpre-etal-2021-entity, liu-etal-2024-lost}. While Representation Engineering \citep{zou2025representationengineeringtopdownapproach} demonstrates the utility of top-down linear control, the geometric interpretation of hallucinations as orthogonal interference remains under-explored. DCO addresses this by reformulating factual consistency—evaluated via metrics such as FactKB \citep{feng-etal-2023-factkb}—as the filtering of orthogonal noise. By integrating challenges identified in summarization benchmarks like XSum \citep{narayan-etal-2018-dont} into this framework, we provide a unified geometric approach that preserves model integrity while suppressing noise, offering a mechanistic alternative to probability-based corrections.

\paragraph{Large Language Model Development and Downstream Applications.}
The rapid scaling of LLMs, as exemplified by bilingual model series such as TeleChat \citep{he2024telechattechnicalreport, wang-etal-2024-telechat} and their successors TeleChat2 and TeleChat2.5 \citep{wang2025technicalreporttelechat2telechat25}, as well as the Mixture-of-Experts architecture TeleChat3-MoE \citep{liu2025trainingreporttelechat3moe}, highlights the increasing complexity and diversity of deployed language systems. Similarly, the Tele-FLM series \citep{li2024teleflmtechnicalreport, li202452b1tlessonslearned} demonstrates the empirical lessons learned when scaling from tens to hundreds of billions of parameters. Across these architectures, hallucination persists as a critical reliability bottleneck. This challenge is particularly pronounced in high-stakes downstream applications, including mathematical reasoning \citep{zhao2025enhancing}, structured table reasoning \citep{xiong2025tablereasoneradvancingtablereasoning}, process-reward-guided structured data construction \citep{xing-etal-2025-llmsr}, and universal information extraction via reinforcement learning \citep{li2025mruiemultiperspectivereasoningreinforcement}, where factual deviations carry direct consequences. Furthermore, model compression research such as Mosaic Pruning for Mixture-of-Experts models \citep{hu2025mosaicpruninghierarchicalframework} underscores the need to maintain representational integrity under efficiency constraints—a concern closely aligned with hallucination-free generation. Our work addresses this fundamental reliability gap through a training-free geometric intervention, providing a principled solution broadly applicable across the aforementioned deployment contexts.

\section{Methodology}

\begin{figure*}[t] 
	\centering
	\includegraphics[width=1.0\textwidth]{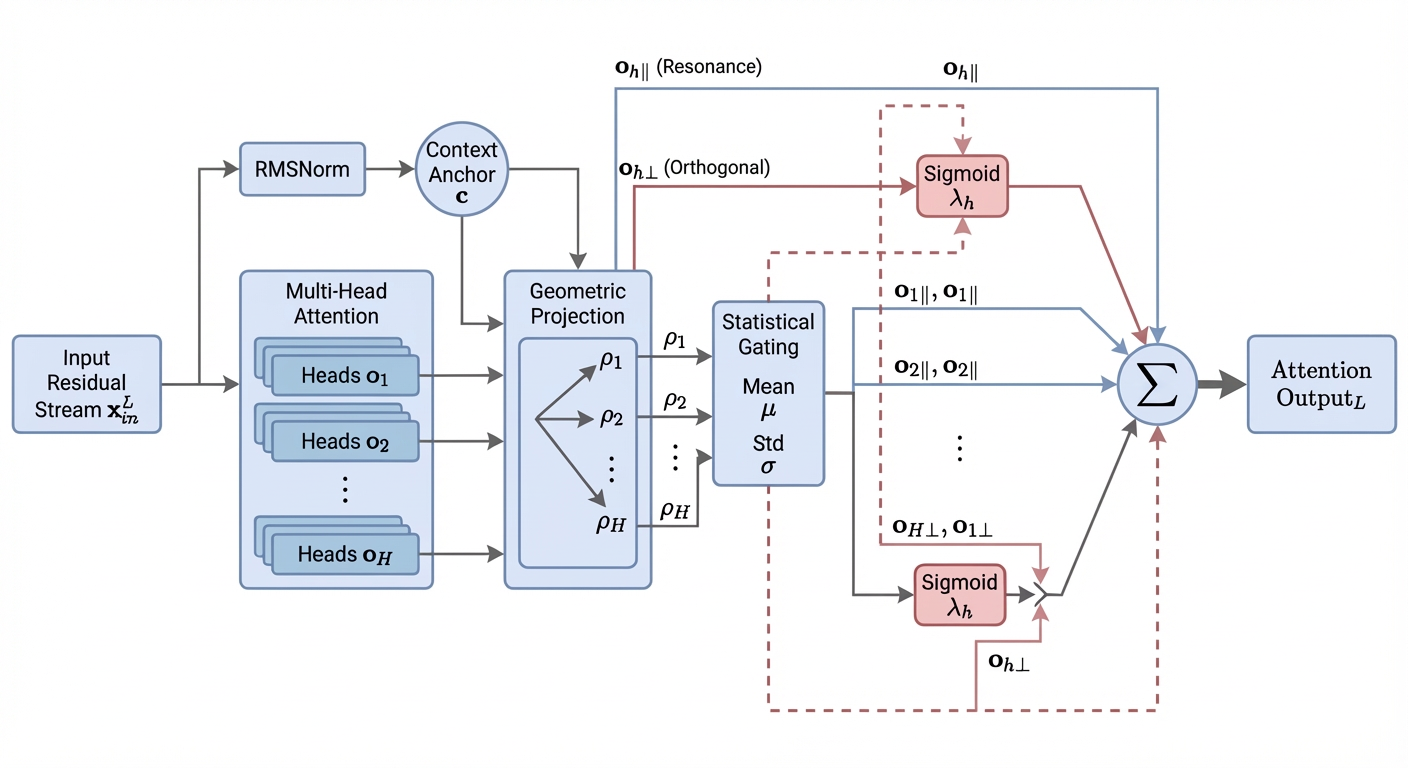}
	\caption{\textbf{Mechanistic pipeline of Dynamic Contextual Orthogonalization (DCO).} 
		The process begins by deriving a context anchor $\mathbf{c}$ from the input residual stream $\mathbf{x}_{in}^L$. Each attention head output $\mathbf{o}_h$ is then orthogonally decomposed into a \textbf{context-aligned component} $\mathbf{o}_{h\parallel}$ and an orthogonal component $\mathbf{o}_{h\perp}$. 
		The aligned path bypasses intervention to preserve logical continuity, while the orthogonal path is modulated by a dynamic suppression coefficient $\lambda_h$, computed via layer-wise Z-score statistics ($\mu, \sigma$) of the orthogonality metric $\rho$. This ensures that only outlier heads—introducing divergent noise relative to the semantic manifold—are attenuated.}
	\label{fig:pipeline}
\end{figure*}

The computational architecture of the DCO intervention is illustrated in \textbf{Figure \ref{fig:pipeline}}. The mechanism of Dynamic Contextual Orthogonalization (DCO) functions by applying linear projection operators to constrain the state evolution of latent representations during inference. This operation targets the output of the Multi-Head Attention (MHA) module in layer $L$, where vectors produced by individual attention heads are orthogonally decomposed to identify and attenuate components that exhibit statistically significant deviation from the contextual semantic manifold.

\subsection{Context Anchor Construction}

The initial phase involves constructing a context anchor, $\mathbf{c}$, which quantitatively defines the direction of semantic consistency at the current logical step. We posit that the input residual stream $\mathbf{x}_{in}^{L} \in \mathbb{R}^{d_{model}}$ at layer $L$ represents the accumulated semantic consensus derived from the context processed up to layer $L-1$. To focus on the directional alignment rather than magnitude, we apply Root Mean Square Normalization (RMSNorm) to the residual vector. The normalized context anchor is defined as:
\begin{equation}
	\mathbf{c} = \frac{\text{RMSNorm}(\mathbf{x}_{in}^{L})}{\|\text{RMSNorm}(\mathbf{x}_{in}^{L})\|_2 + \epsilon}
\end{equation}
where $\epsilon$ is a small stability constant. Geometrically, $\mathbf{c}$ serves as the reference unit vector on the hypersphere of the latent space, delineating the principal direction of the ongoing generative process.

\subsection{Subspace-based Orthogonal Decomposition}

The second phase executes a fine-grained decomposition of attention head outputs. Let $H$ denote the number of attention heads in layer $L$. For the $h$-th attention head, let $\mathbf{o}_h \in \mathbb{R}^{d_{model}}$ be its output vector. Following the principles of linear projection, we partition $\mathbf{o}_h$ into a \textbf{context-aligned component} $\mathbf{o}_{h\parallel}$ and a \textbf{context-orthogonal component} $\mathbf{o}_{h\perp}$.

The context-aligned component, representing information congruent with the context anchor, is computed via projection:
\begin{equation}
	\mathbf{o}_{h\parallel} = (\mathbf{o}_h \cdot \mathbf{c}) \cdot \mathbf{c}
\end{equation}
The orthogonal component is defined as the vector rejection:
\begin{equation}
	\mathbf{o}_{h\perp} = \mathbf{o}_h - \mathbf{o}_{h\parallel}
\end{equation}
To quantify the divergence of each head relative to the context, we define the orthogonality metric $\rho_h$ as the ratio of the orthogonal component's magnitude to the total magnitude:
\begin{equation}
	\rho_h = \frac{\|\mathbf{o}_{h\perp}\|_2}{\|\mathbf{o}_h\|_2 + \epsilon}
\end{equation}
As $\rho_h \to 0$, the head reinforces the existing context; as $\rho_h \to 1$, the head introduces information linearly independent of the established semantic manifold.

\subsection{Z-Score Based Dynamic Suppression}

To achieve adaptive intervention, DCO relies on the statistical distribution of $\rho$ within the current layer rather than static thresholds. We observe that in high-dimensional semantic spaces, valid information diversification and hallucinatory noise often overlap in absolute magnitude but differ in their relative statistical deviation.

\begin{figure}[ht]
	\centering
	\includegraphics[width=0.95\columnwidth]{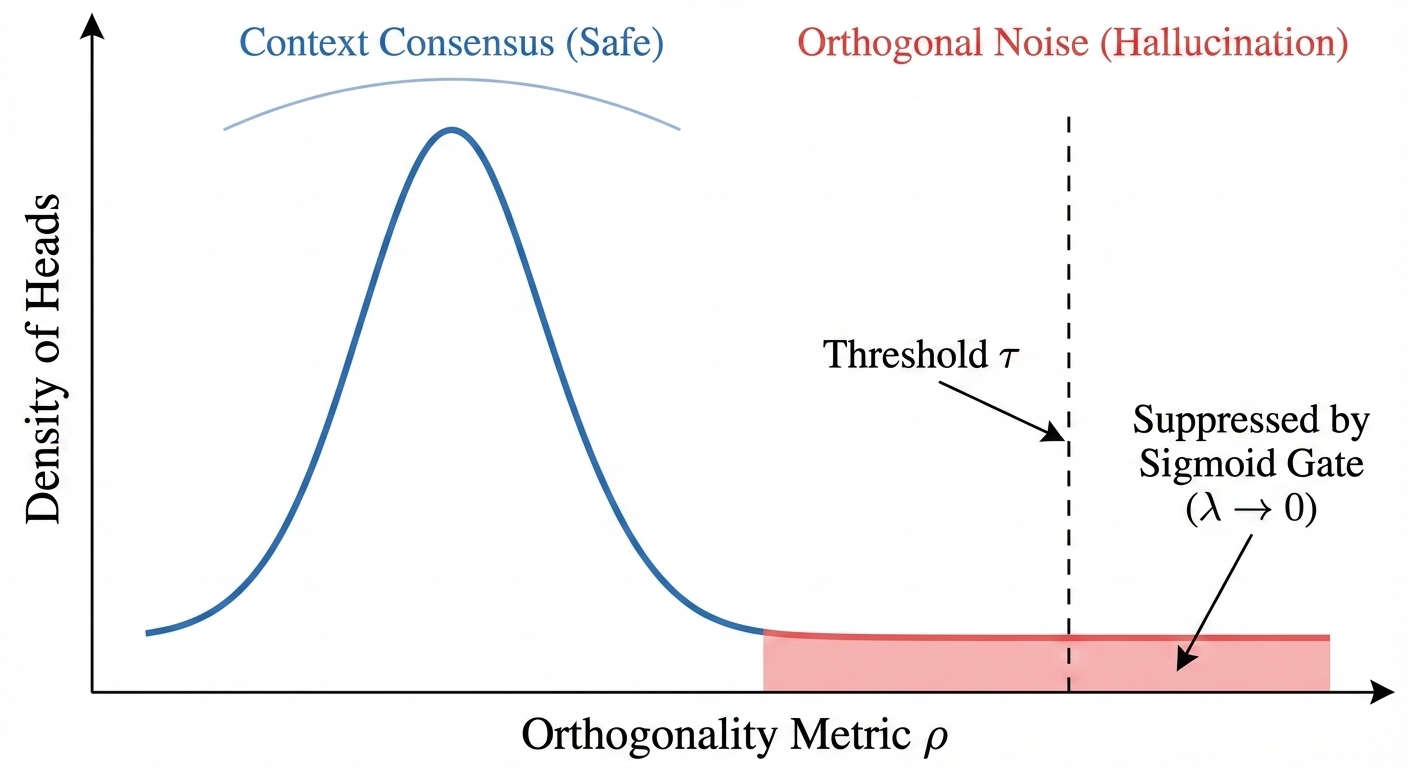}
	\caption{\textbf{Schematic of the Z-Score dynamic suppression mechanism.} 
		We model the distribution of orthogonality metrics $\rho$ across attention heads. 
		Most heads cluster within the context-aligned region (blue peak), while hallucinating heads manifest as statistical outliers in the high-orthogonality tail (red region). 
		DCO dynamically sets a threshold $\tau$ derived from layer-wise statistics to selectively suppress these divergent components via a Sigmoid gate.}
	\label{fig:distribution}
\end{figure}

We first calculate the mean $\mu_{\rho}$ and standard deviation $\sigma_{\rho}$ across all $H$ heads:
\begin{equation}
	\mu_{\rho} = \frac{1}{H} \sum_{h=1}^{H} \rho_h, \quad \sigma_{\rho} = \sqrt{\frac{1}{H} \sum_{h=1}^{H} (\rho_h - \mu_{\rho})^2}
\end{equation}
The standard score (Z-Score) $z_h$ for each head is derived to measure relative deviation:
\begin{equation}
	z_h = \frac{\rho_h - \mu_{\rho}}{\sigma_{\rho} + \epsilon}
\end{equation}
As illustrated in Figure \ref{fig:distribution}, we hypothesize that hallucinating heads manifest as statistical outliers in the long tail of the orthogonality distribution. To softly attenuate these outliers while preserving gradient continuity, we employ a non-linear Sigmoid gating function to determine the suppression coefficient $\lambda_h \in (0, 1]$:
\begin{equation}
	\lambda_h = \frac{1}{1 + \exp(\beta \cdot (z_h - \tau))}
\end{equation}

Here, $\tau$ represents the tolerance threshold in standard deviation units, and $\beta$ is the temperature coefficient controlling the steepness of the suppression curve. The final reconstructed output of the attention module at layer $L$ is the summation of the intervened head vectors:
\begin{equation}
	\text{AttentionOutput}_L = \sum_{h=1}^{H} (\mathbf{o}_{h\parallel} + \lambda_h \cdot \mathbf{o}_{h\perp})
\end{equation}
This formulation ensures that heads exhibiting normative behavior ($z_h < \tau$) retain their orthogonal contributions ($\lambda_h \approx 1$), thereby preserving necessary semantic diversity.

\subsection{Implementation Strategy}

DCO is deployed using a layer-selective strategy, targeting the middle-to-late layers defined by $L \in [L_{start}, L_{end}]$. This design is grounded in mechanistic findings that early layers ($L < L_{start}$) are critical for parsing local syntax and often exhibit high intrinsic orthogonality required for structural composition.

The computational overhead of DCO is negligible. For a model with hidden dimension $d_{model}$ and $H$ heads, the complexity is $O(L \cdot H \cdot d_{model})$. Since $d_{model}$ is constant and $H \cdot d_{model}$ scales linearly with the model size, this operation is significantly more efficient than the quadratic attention mechanism ($O(N^2 \cdot d_{model})$), ensuring minimal latency increase in production environments.

\begin{figure}[ht] 
	\centering
	\includegraphics[width=0.95\columnwidth]{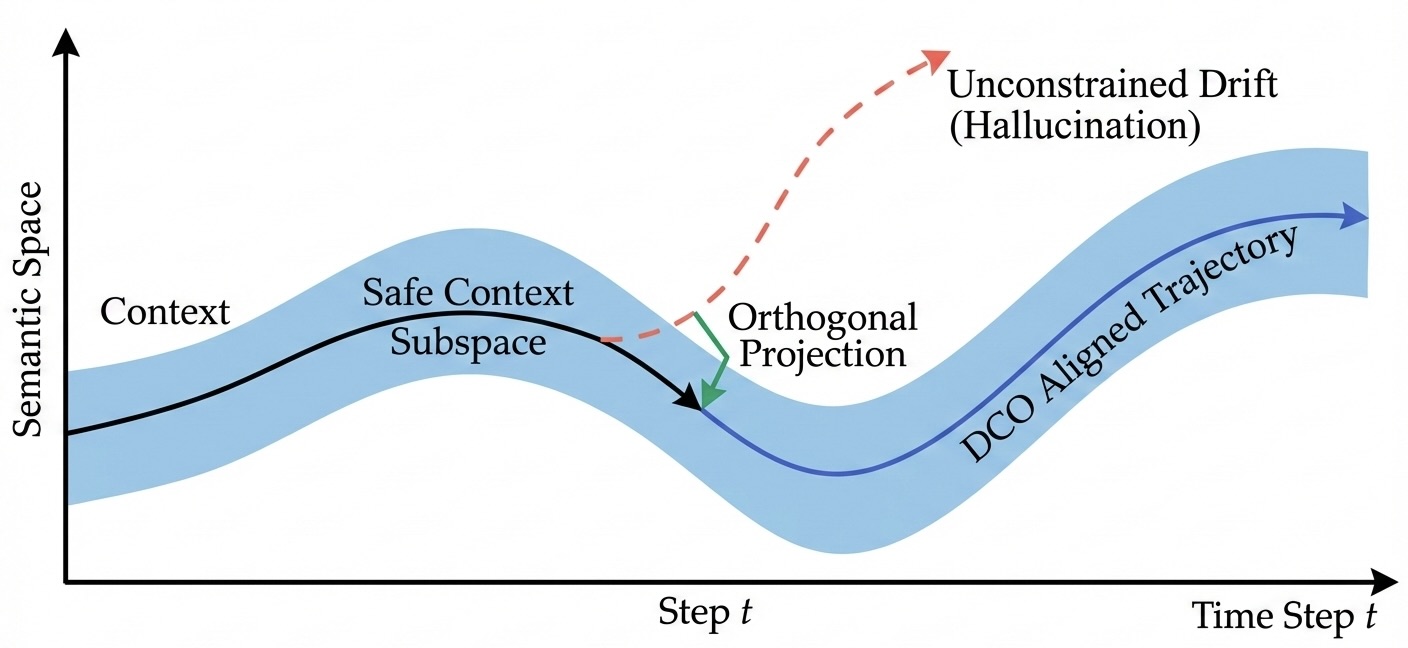} 
	\caption{\textbf{Visualization of latent manifold dynamics.} 
		The blue ribbon represents the \textit{Context Manifold} $\mathcal{M}$. 
		Without intervention, the generative trajectory (Red dashed line) suffers from \textit{orthogonal drift}, diverging into hallucination. 
		DCO (Blue solid line) applies continuous orthogonal projection (Green arrow) at each step $t$, constraining the latent state evolution within the safe semantic subspace.}
	\label{fig:trajectory}
\end{figure}

\section{Results and Analysis}

We evaluate the efficacy of Dynamic Contextual Orthogonalization (DCO) across a spectrum of benchmarks to assess its impact on faithfulness, factuality, and reasoning stability. All experiments are conducted using the Llama-3-8B-Instruct and Llama-3-70B-Instruct models \citep{grattafiori2024llama3herdmodels}.

We evaluate contextual faithfulness and instruction following using IFEval \citep{Zhou2023InstructionFollowingEF}, XSum, and NQ-Swap. For factuality and knowledge retention, we employ TruthfulQA \citep{lin2022truthfulqameasuringmodelsmimic}, TriviaQA \citep{joshi2017triviaqalargescaledistantly}, and Natural Questions (NQ-Open) \citep{kwiatkowski-etal-2019-natural}. Complex reasoning stability is assessed via MuSiQue \citep{trivedi2022musiquemultihopquestionssinglehop}.

\begin{table*}[t]
	\centering
	\small 
	\renewcommand{\arraystretch}{0.9} 
	\caption{Performance comparison on faithfulness evaluation tasks. We report ROUGE-L (R-L), BERTScore \citep{zhang2020bertscoreevaluatingtextgeneration}, and FactKB \citep{feng-etal-2023-factkb} scores. Note that \textbf{NQ-Open} here is evaluated in an \textbf{Open-Book} setting (with retrieved context).}
	\label{tab:faithfulness}
	\resizebox{0.95\textwidth}{!}{%
		\begin{tabular}{llccccccccc}
			\toprule
			\multirow{2}{*}{\textbf{Model}} & \multirow{2}{*}{\textbf{Method}} & \multicolumn{3}{c}{\textbf{XSum}} & \multicolumn{2}{c}{\textbf{MemoTrap}} & \multicolumn{2}{c}{\textbf{IFEval}} & \textbf{NQ-Open} & \textbf{NQ-Swap} \\
			\cmidrule(lr){3-5} \cmidrule(lr){6-7} \cmidrule(lr){8-9} \cmidrule(lr){10-10} \cmidrule(lr){11-11}
			& & \textbf{R-L} & \textbf{BS-F1} & \textbf{FactKB} & \textbf{Macro} & \textbf{Micro} & \textbf{Prompt} & \textbf{Inst.} & \textbf{(Open)} & \textbf{EM} \\
			\midrule
			\textbf{Llama-3-8B} 
			& Greedy & 19.90 & 67.23 & 47.61 & 65.86 & 64.40 & \underline{70.24} & 78.30 & 69.68 & 60.62 \\
			& ITI & 13.25 & 59.96 & 34.35 & 62.65 & 58.96 & 52.31 & 63.19 & 56.16 & 51.08 \\
			& CAD & 18.82 & 67.20 & \textbf{67.16} & - & - & - & - & 69.83 & \textbf{74.21} \\
			& DoLa (low) & 19.82 & 67.19 & 47.21 & 65.27 & 63.69 & 69.69 & 78.18 & 69.68 & 60.77 \\
			& DoLa (high) & \underline{19.92} & 67.34 & 48.49 & 64.85 & 63.17 & \underline{70.24} & \underline{78.66} & 69.49 & 60.98 \\
			& AD & 19.79 & 67.31 & 48.49 & 65.38 & 64.28 & 67.65 & 76.26 & 68.93 & 60.51 \\
			& DeCoRe (static) & 19.87 & \underline{67.83} & \underline{64.07} & \textbf{69.53} & \textbf{69.20} & 69.13 & 78.06 & \underline{70.62} & \underline{64.43} \\
			\rowcolor{gray!10} & \textbf{DCO (Ours)} & \textbf{20.39} & \textbf{67.86} & 56.70 & \underline{67.06} & \underline{65.30} & \textbf{74.68} & \textbf{81.41} & \textbf{74.95} & 62.96 \\
			\midrule
			\textbf{Llama-3-70B} 
			& Greedy & 22.41 & 69.77 & 61.32 & 68.47 & 66.52 & 77.45 & 84.41 & 71.07 & 76.11 \\
			& ITI & 21.64 & 69.46 & 61.33 & \underline{71.24} & \underline{68.73} & 76.71 & 83.69 & 71.90 & 74.76 \\
			& CD & \textbf{22.71} & \textbf{69.99} & 54.73 & 69.27 & 67.55 & 71.72 & 79.74 & 65.80 & 68.37 \\
			& CAD & 21.45 & 69.28 & \textbf{65.61} & - & - & - & - & 71.83 & \textbf{84.70} \\
			& DoLa (low) & 22.46 & 69.80 & 61.11 & 67.99 & 65.93 & 77.08 & 84.29 & 71.07 & 75.98 \\
			& DoLa (high) & 22.43 & \underline{69.93} & 59.99 & 67.92 & 65.81 & 78.00 & 84.65 & 70.40 & 75.26 \\
			& AD & \underline{22.49} & 69.91 & 60.57 & 67.51 & 66.44 & 76.89 & 84.41 & 71.15 & 74.02 \\
			& DeCoRe (static) & 21.94 & 69.35 & \underline{64.88} & \textbf{71.96} & \textbf{71.41} & \underline{78.56} & \underline{84.89} & \underline{72.51} & \underline{79.06} \\
			\rowcolor{gray!10} & \textbf{DCO (Ours)} & 22.40 & 69.74 & 60.87 & 69.17 & 67.19 & \textbf{85.77} & \textbf{90.29} & \textbf{79.51} & 75.33 \\
			\bottomrule
		\end{tabular}%
	}
\end{table*}

\subsection{Experimental Configuration}

To maintain intervention stability across diverse semantic landscapes, we adopt a calibration strategy based on \textit{semantic density} and \textit{reasoning depth}. The intervention is primarily applied to the middle-to-late layers of the residual stream, aligning with mechanistic findings that these layers govern factual retrieval and integration. 

The hyperparameters—tolerance threshold $\tau$ and temperature coefficient $\beta$—are configured to reflect the geometric constraints of specific tasks. For counterfactual tasks requiring strict adherence to provided evidence (e.g., NQ-Swap), we apply tighter projection constraints to filter parametric bias. Conversely, for multi-hop reasoning (e.g., MuSiQue), constraints are relaxed to accommodate the necessary semantic expansion required for logic chaining. This task-dependent configuration ensures that DCO adapts to the specific manifold geometry of downstream applications.

\begin{table*}[h]
	\centering
	\small
	\renewcommand{\arraystretch}{0.9}
	\caption{Performance comparison on factuality benchmarks. We focus on TruthfulQA (MC), TriviaQA, and NQ-Open in the closed-book setting.}
	\label{tab:factuality}
	\resizebox{0.72\textwidth}{!}{
		\begin{tabular}{llccccc}
			\toprule
			\multirow{2}{*}{\textbf{Model}} & \multirow{2}{*}{\textbf{Method}} & \multicolumn{3}{c}{\textbf{TruthfulQA (MC)}} & \textbf{TriviaQA} & \textbf{NQ-Open} \\
			\cmidrule(lr){3-5} \cmidrule(lr){6-6} \cmidrule(lr){7-7}
			& & \textbf{MC1} & \textbf{MC2} & \textbf{MC3} & \textbf{EM} & \textbf{(Closed)} \\
			\midrule
			\textbf{Llama-3-8B} 
			& Greedy & 39.41 & 55.69 & 30.31 & 56.58 & 29.04 \\
			& ITI & \textbf{43.70} & \textbf{62.78} & \textbf{34.91} & 48.41 & 22.07 \\
			& DoLa (low) & 39.05 & 55.65 & 30.06 & 56.63 & 29.15 \\
			& DoLa (high) & 38.68 & 55.64 & 30.19 & 56.50 & 29.19 \\
			& AD & 31.21 & 55.30 & 28.28 & 54.93 & 28.32 \\
			& DeCoRe (static) & 38.68 & 55.74 & 29.80 & \underline{56.93} & \underline{29.42} \\
			\rowcolor{gray!10} & \textbf{DCO (Ours)} & \underline{40.39} & \underline{58.68} & \underline{32.21} & \textbf{57.33} & \textbf{30.92} \\
			\midrule
			\textbf{Llama-3-70B} 
			& Greedy & 49.57 & 70.60 & 37.85 & 74.77 & 40.08 \\
			& ITI & 48.96 & 67.04 & 37.27 & 73.54 & 38.57 \\
			& CD & \textbf{57.77} & \textbf{76.65} & \textbf{47.08} & 72.83 & 36.23 \\
			& DoLa (low) & 49.45 & 70.58 & 37.75 & 74.74 & 40.08 \\
			& DoLa (high) & 49.69 & 70.88 & 38.01 & 73.96 & 39.59 \\
			& AD & 42.23 & 67.56 & 35.37 & 74.14 & 40.23 \\
			& DeCoRe (static) & \underline{51.29} & \underline{72.02} & \underline{40.24} & \underline{74.79} & \underline{40.41} \\
			\rowcolor{gray!10} & \textbf{DCO (Ours)} & 48.23 & 69.56 & 36.73 & \textbf{76.12} & \textbf{43.62} \\
			\bottomrule
		\end{tabular}
	}
\end{table*}
\subsection{Performance on Contextual Faithfulness}

Table~\ref{tab:faithfulness} details the performance of DCO on benchmarks evaluating adherence to external contexts and instructions. DCO demonstrates statistically significant improvements over baselines, particularly in instruction following and context-grounded retrieval tasks.

On the IFEval benchmark, DCO achieves state-of-the-art results across model scales. For Llama-3-70B, DCO elevates the Prompt-level strict accuracy to 85.77\% and Instruction-level accuracy to 90.29\%, representing a substantial margin over the greedy baseline of 77.45\% and 84.41\%, respectively, and exceeding existing intervention methods such as DeCoRe. Similarly, in the NQ-Open Open-Book setting, DCO outperforms all baselines, reaching 79.51\% on the 70B model. These results suggest that enforcing geometric alignment within the residual stream effectively enhances the capacity of the model to prioritize contextual evidence over internal parametric noise.

Regarding the XSum summarization task, DCO achieves the highest ROUGE-L of 20.39 on the 8B model. Crucially, when compared to Inference-Time Intervention (ITI), which utilizes static steering vectors, DCO exhibits superior robustness. While ITI suffers a severe performance regression, dropping to 13.25 R-L, DCO maintains high linguistic quality. This disparity validates the hypothesis that dynamic, context-dependent orthogonalization is essential for maintaining the integrity of the latent manifold.

In the NQ-Swap counterfactual task, while Context-Aware Decoding (CAD) retains the highest EM scores through logit-level contrastive amplification, DCO serves as a computationally efficient alternative. By achieving competitive alignment of 75.33\% EM on the 70B model via a single-forward projection, DCO avoids the dual-decoding latency overhead inherent in contrastive methods while still effectively suppressing parametric interference.

\begin{table*}[t]
	\centering
	\small
	\renewcommand{\arraystretch}{1.0}
	\caption{Multi-hop reasoning performance on MuSiQue.}
	\label{tab:musique}
	\resizebox{0.75\textwidth}{!}{%
		\begin{tabular}{llcccc}
			\toprule
			\multirow{2}{*}{\textbf{Model}} & \multirow{2}{*}{\textbf{Method}} & \multicolumn{2}{c}{\textbf{Without CoT}} & \multicolumn{2}{c}{\textbf{With CoT}} \\
			\cmidrule(lr){3-4} \cmidrule(lr){5-6}
			& & \textbf{Closed Book} & \textbf{Open Book} & \textbf{Closed Book} & \textbf{Open Book} \\
			\midrule
			\textbf{Llama-3-8B} 
			& Greedy & \underline{7.41} & 58.83 & 14.61 & 69.84 \\
			& ITI & 4.01 & 45.84 & 4.18 & 38.31 \\
			& CAD & - & 57.88 & - & \textbf{73.02} \\
			& DoLa & 7.24 & \underline{59.08} & \textbf{14.94} & 69.92 \\
			& AD & 6.99 & 58.63 & 14.40 & 69.92 \\
			& DeCoRe (static) & \textbf{7.90} & \textbf{61.23} & \underline{14.69} & \underline{72.49} \\
			\rowcolor{gray!10} & \textbf{DCO (Ours)} & \underline{7.41} & 58.26 & 14.56 & 69.01 \\
			\midrule
			\textbf{Llama-3-70B} 
			& Greedy & \underline{11.79} & 68.56 & 20.15 & 74.43 \\
			& ITI & 10.88 & 68.14 & 20.44 & 74.27 \\
			& CD & 10.92 & 66.61 & 17.17 & 71.70 \\
			& CAD & - & 68.64 & - & 74.02 \\
			& DoLa & 11.42 & \underline{68.68} & 20.15 & \underline{74.64} \\
			& AD & 11.38 & 68.14 & 20.23 & 74.27 \\
			& DeCoRe (static) & \underline{11.79} & \textbf{69.76} & \underline{20.60} & \textbf{75.05} \\
			\rowcolor{gray!10} & \textbf{DCO (Ours)} & \textbf{11.96} & 67.89 & \textbf{21.14} & 74.27 \\
			\bottomrule
		\end{tabular}%
	}
\end{table*}

\subsection{Pareto Frontier: Knowledge Preservation}

An essential consideration in hallucination mitigation is the preservation of the model's general capabilities, as aggressive intervention often leads to a degradation of parametric knowledge. Table~\ref{tab:factuality} illustrates this performance trade-off. While methods such as ITI and CD report higher scores on TruthfulQA, they frequently cause a significant performance regression in factual retrieval. For instance, on the Llama-3-8B model, ITI results in a decrease in TriviaQA accuracy from 56.58\% to 48.41\%, indicating that static interventions may inadvertently suppress valid factual retrieval pathways alongside hallucinatory signals.

In contrast, DCO optimizes the Pareto frontier by maintaining or enhancing performance on knowledge-intensive benchmarks. On the Llama-3-70B model, DCO achieves the highest performance among the evaluated methods for both TriviaQA (76.12\%) and NQ-Open Closed-book (43.62\%) settings. This robustness is attributed to the adaptive Z-score statistical filtering mechanism. By dynamically defining suppression thresholds based on layer-wise distributions, DCO distinguishes between the normative statistical signals of factual retrieval and the outlier components characteristic of hallucinations. These results provide empirical evidence that dynamic orthogonalization effectively minimizes interference with standard knowledge retrieval processes while maintaining competitive truthfulness scores, such as the 40.39\% MC1 achieved on the 8B model.

\subsection{Reasoning Stability in Complex Chains}

Table~\ref{tab:musique} evaluates the performance of various intervention methods on the MuSiQue multi-hop reasoning dataset. On the Llama-3-70B model, DCO achieves the highest performance in the Closed-book setting when utilized with Chain-of-Thought (CoT), reaching an Exact Match (EM) score of 21.14\%. This represents an improvement over both the greedy baseline (20.15\%) and head-level intervention methods such as DeCoRe (20.60\%).

The observed improvement is posited to stem from the mitigation of compounding deviations within the residual stream during multi-step inference. In complex reasoning chains, minor orthogonal components introduced in initial logical steps may accumulate, leading to semantic drift where the latent trajectory diverges from the factual manifold. By enforcing manifold alignment at each layer, DCO attenuates these divergent components before they propagate through the network. Furthermore, the competitive performance of DCO relative to head-masking approaches like DeCoRe suggests that the errors associated with multi-hop reasoning may be distributive in nature. Consequently, these errors appear to be more effectively addressed via global statistical constraints on the residual stream rather than the isolation of individual attention heads. While DoLa and DeCoRe maintain a slight advantage in certain Open-book scenarios, DCO demonstrates robust stability in internalizing reasoning paths without external evidence.

\begin{figure}[h]
	\centering
	\includegraphics[width=0.85\columnwidth]{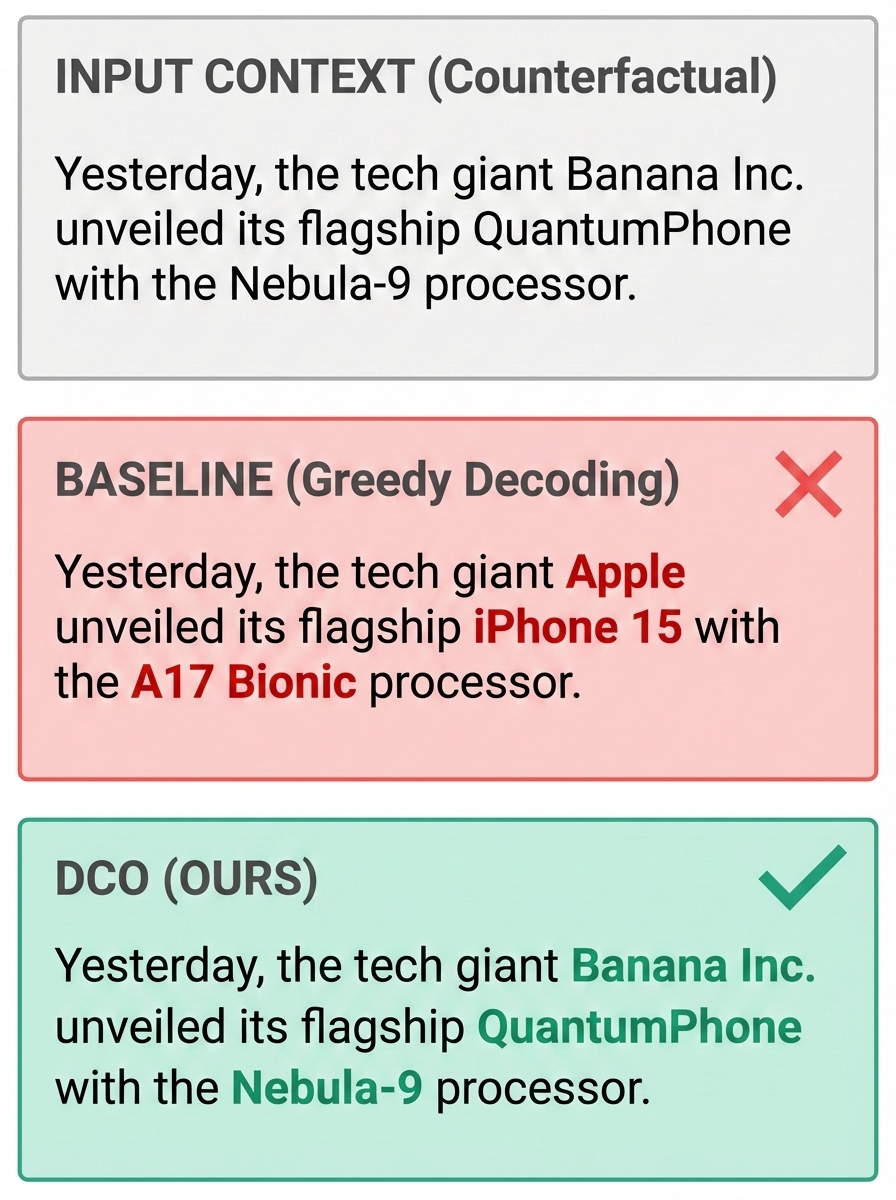}
	\caption{\textbf{Qualitative comparison under knowledge conflict.} 
		Top: The input context contains counterfactual entities (``Banana Inc.''). 
		Middle: The baseline model (Red) ignores the context and hallucinates real-world entities (``Apple'', ``iPhone'') due to parametric memory bias. 
		Bottom: DCO (Green) successfully suppresses this orthogonal noise, aligning the generation with the counterfactual context.}
	\label{fig:case_study}
\end{figure}

\subsection{Qualitative Analysis}
\label{sec:qualitative}

To examine the microscopic mechanism of DCO, we analyze a "knowledge conflict" scenario in Figure~\ref{fig:case_study}, where fictitious context contradicts the model's parametric priors. In the baseline model, specific attention heads dominated by pre-trained associations generate real-world entities (e.g., "Apple"), effectively ignoring the provided context. Geometrically, these heads produce output vectors that are linearly independent of the context anchor $\mathbf{c}$, representing orthogonal drift from the semantic manifold. 

DCO identifies these divergent components as statistical outliers in the Z-score distribution of the orthogonality metric $\rho$. By applying the dynamic gating mechanism, DCO attenuates these anomalous orthogonal vectors while preserving components collinear with the context anchor. This operation effectively projects the divergent generative trajectory back onto the context manifold $\mathcal{M}$ without disrupting syntactic coherence. These findings provide empirical evidence for the hypothesis of hallucinations as orthogonal noise and demonstrate DCO's efficacy in enforcing contextual alignment under conflict.

\section{Conclusion}

In this work, we have presented Dynamic Contextual Orthogonalization (DCO), a framework that operationalizes the geometric hypothesis of hallucinations as \textit{orthogonal noise injection}. By formalizing factual deviations as vector components that diverge from the contextual semantic manifold, we established a mechanistic basis for intervention. DCO utilizes the instantaneous residual stream as a dynamic anchor and employs a layer-wise Z-Score suppression mechanism. This design addresses the rigidity of static steering vectors by adapting to the local statistical distribution of attention outputs, thereby distinguishing between necessary semantic diversification and anomalous noise.

Empirical validation across the Llama-3 family confirms that DCO effectively resolves the persistent trade-off between contextual faithfulness and parametric knowledge retention. Unlike traditional decoding strategies that often degrade general reasoning capabilities to enforce context adherence, DCO's outlier-based attenuation preserves the model's underlying logical structure while significantly enhancing performance on faithfulness benchmarks such as XSum and IFEval. The method's robustness on TriviaQA and MuSiQue further validates that enforcing geometric constraints is a non-destructive pathway to reliable generation.

In summary, DCO offers a computationally efficient, training-free mechanism for alignment during inference. It underscores the necessity of dynamic, representation-level intervention for suppressing hallucinations in complex semantic spaces. Future research will focus on developing meta-learning frameworks to automate the calibration of geometric constraints, aiming to achieve self-adaptive manifold control that generalizes across varying semantic densities.

\section{Limitations}

While Dynamic Contextual Orthogonalization (DCO) demonstrates significant efficacy in enhancing faithfulness and mitigating hallucinations, we identify specific boundary conditions and design choices inherent to the framework.

The implementation of DCO introduces additional linear projection and statistical normalization operations at each decoding step. While this theoretically increases the floating-point operations (FLOPs) per layer, the added complexity scales linearly with the model dimension ($O(d_{model})$). This is negligible compared to the quadratic complexity of the self-attention mechanism ($O(N^2)$) inherent to Transformer architectures. Furthermore, unlike contrastive decoding paradigms (e.g., CAD, DoLa) which necessitate multiple forward passes or parallel decoding of different model states to compute logit differences, DCO operates strictly within a single inference pass. Consequently, in production environments prioritizing throughput and latency, DCO maintains a distinct efficiency advantage over multi-pass intervention strategies.

The geometric definition of the semantic manifold in DCO relies on the extraction of a context anchor from the input residual stream. This design is explicitly optimized for context-grounded generation, summarization, and context-grounded reasoning, where an external reference exists. In scenarios entirely devoid of context (pure closed-book generation), the definition of orthogonality becomes less distinct. However, our empirical results on knowledge-intensive benchmarks (e.g., TriviaQA, TruthfulQA) indicate that the dynamic Z-score filtering mechanism is robust enough to distinguish between valid parametric retrieval and stochastic noise even in lower-context settings. This suggests that DCO successfully mitigates the trade-off between hallucination suppression and general capability degradation, a common failure mode in static intervention methods.

Our framework is predicated on the Linear Representation Hypothesis, positing that semantic deviations manifest as orthogonal vectors in the residual stream. While the underlying neural architecture involves non-linear activation functions, the effectiveness of DCO supports the utility of linear approximations for manipulating high-level semantic features. This geometric simplification provides a mechanistically interpretable control method, offering a more transparent alternative to black-box modifications of output probability distributions.

\bibliography{custom}

\appendix

\end{document}